\documentclass{article}

    \usepackage[preprint]{neurips_2025}

\usepackage[utf8]{inputenc} %
\usepackage[T1]{fontenc}    %
\usepackage{hyperref}       %
\usepackage{url}            %
\usepackage{booktabs}       %
\usepackage{amsfonts}       %
\usepackage{nicefrac}       %
\usepackage{microtype}      %
\usepackage{xcolor}         %

\usepackage{wrapfig}
\usepackage{amssymb}%
\usepackage{pifont}%
\usepackage{multirow}
\usepackage{graphicx}

\usepackage{multirow}
\usepackage{amsmath}
\usepackage{enumitem}
\usepackage{tcolorbox}
\tcbuselibrary{listingsutf8}

\newcommand{\xmark}{\ding{55}}%

\newcommand{\Mat}{\boldsymbol}
\newcommand{\Set}{\mathcal}

\newcommand{\real}{\mathbb{R}}

\newcommand{\xtodo}[1]{#1}

\newcommand{\ours}{MMHU}
\newcommand{\bfours}{\textbf{\ours}}
\newcommand{\paragraphVspace}

\usepackage{lipsum}

\definecolor{rowcolor}{RGB}{209, 233, 246}

\title{MMHU:
A Massive-Scale Multimodal Benchmark for Human Behavior Understanding 
}

\author{
  \textbf{Renjie Li}$^{1}$\thanks{Equal contribution},
  \textbf{Ruijie Ye}$^{2*}$,
  \textbf{Mingyang Wu}$^{1*}$,
  \textbf{Hao Frank Yang}$^{3}$,\\
  \textbf{Zhiwen Fan}$^{4}$,
  \textbf{Hezhen Hu}$^{4}$\footnotemark[2],
  \textbf{Zhengzhong Tu}$^{1}$\footnotemark[2]\thanks{Corresponding authors.} \\
  $^1$Texas A\&M University \quad
  $^2$Brown University \quad
  $^3$Johns Hopkins University \quad
  $^4$UT Austin\quad
  \vspace{-9mm}
}

\begin{document}

\maketitle

\vspace{0.3cm}
\centerline{\qquad \textbf{\color{magenta} Project Page}: \url{https://MMHU-Benchmark.github.io}}

\begin{figure*}[ht]
    \centering
 \includegraphics[width=0.99\linewidth]{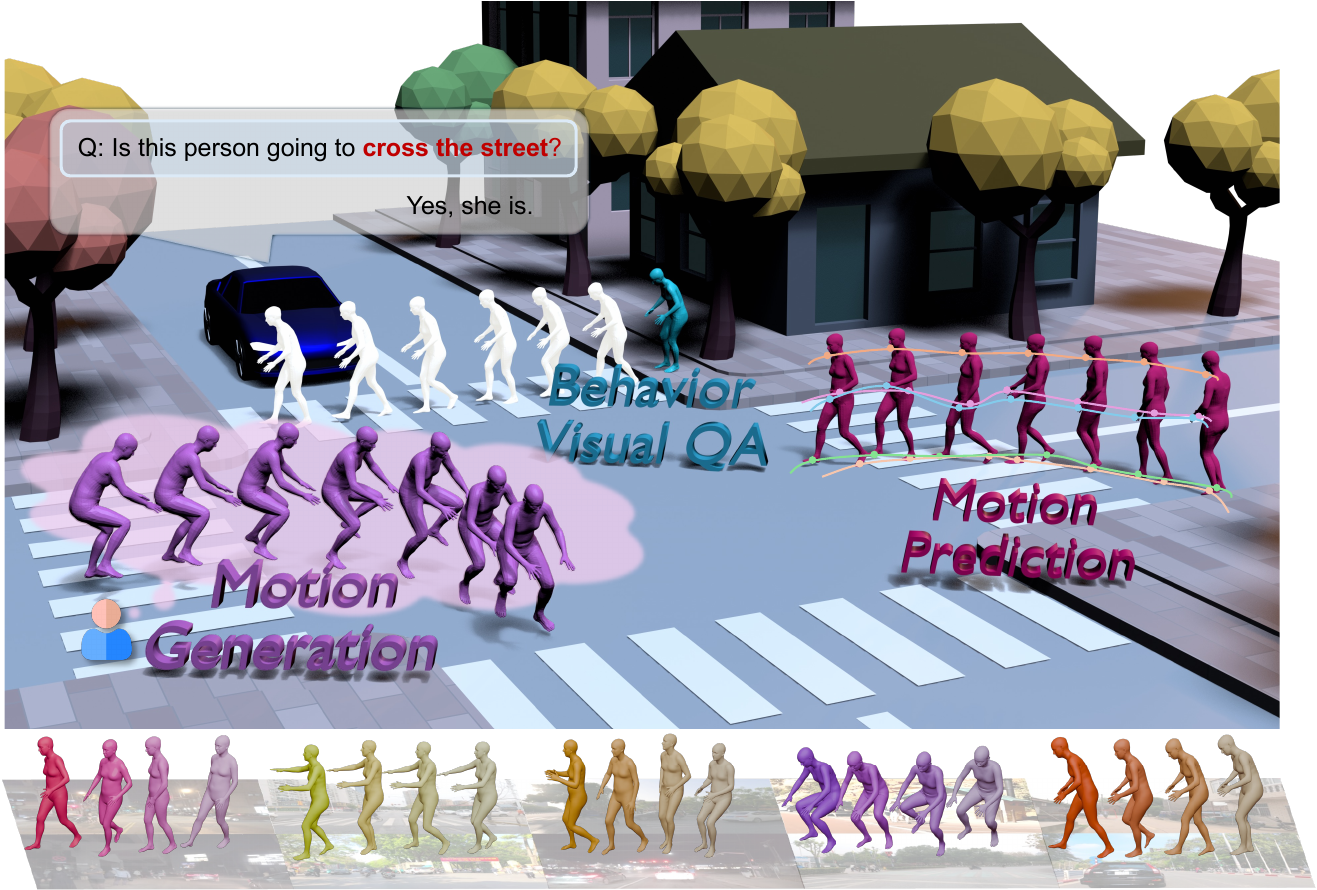}
           \caption{We propose \bfours, a large-scale dataset for human behavior understanding. 
    We collected 57k human instances with diverse behaviors such as playing mobile phone, holding object, or using mobility devices, from diverse scenes such as in the city, school, park, and alley. We provide rich annotations including motion and trajectory, text descriptions for human motions, and recognize the behaviors that are critical to driving safety. 
    }
    \label{fig:teaser}
    
\end{figure*}

\begin{abstract}
Humans are integral components of the transportation ecosystem, and understanding their behaviors is crucial to facilitating the development of safe driving systems. Although recent progress has explored various aspects of human behavior---such as motion, trajectories, and intention---a comprehensive benchmark for evaluating human behavior understanding in autonomous driving remains unavailable.
In this work, we propose \bfours, a large-scale benchmark for human behavior analysis featuring rich annotations, such as human motion and trajectories, text description for human motions, human intention, and critical behavior labels relevant to driving safety. 
Our dataset encompasses 57k human motion clips and 1.73M frames gathered from diverse sources, including established driving datasets such as Waymo, in-the-wild videos from YouTube, and self-collected data. 
A human-in-the-loop annotation pipeline
is developed to generate rich behavior captions. 
We provide a thorough dataset analysis and benchmark multiple tasks—ranging from motion prediction to motion generation and human behavior question answering—thereby offering a broad evaluation suite. Our dataset will be released to promote further human-centric research in this vital area of autonomous driving.

\end{abstract}

\vspace{-3mm}

\section{Introduction}
\label{sec:intro}
\vspace{-2mm}

Humans play an essential role in transportation systems, making the comprehensive understanding of human behaviors---such as motion \cite{xu2023auxiliary}, intention \cite{osman2023tamformer, xie2024gtranspdm, yang2022predicting}, and trajectory \cite{zhang2024incorporating, fang2024behavioral}---critical for developing safe autonomous driving systems. 
To effectively interact with humans, autonomous vehicles must answer human-centric questions, such as “What is the person doing?”, “Is the person going to cross the street?”, and “Where does the person intend to go?”
Failing to accurately comprehend these behaviors could lead to misinterpretations of human intent, potentially resulting in fatal accidents.

While significant efforts have been devoted to understanding individual aspects of human behaviors by investigating human motion, intention, and trajectory, the absence of a unified dataset limits the comprehensive evaluation of algorithms for human behavior understanding, especially in autonomous driving scenarios that mainly account for human safety. Existing driving datasets are typically designed for general driving tasks, such as depth estimation, 2D or 3D object detection, odometry, and semantic segmentation~\cite{sun2020scalability, caesar2020nuscenes, geiger2013vision}, or narrowly designed human-related tasks, such as intention prediction~\cite{rasouli2019pie, bhattacharyya2021euro, rasouli2017they}, motion and trajectory prediction~\cite{von2018recovering}, or motion reconstruction~\cite{wangpedestrian, kim2024text, liu2024learning}.
Moreover, with the emergence of driving-oriented vision-language models (VLMs)~\cite{touvron2023llama,lin2023vila, chen2024longvila, liu2024nvila, liu2023llava, liu2024llavanext, liu2023improvedllava, zhang2024videoinstructiontuningsynthetic, bai2023qwen} , human behavior understanding tasks can now be approached in a more integrated and flexible manner through images and text queries. 
However, existing training data for these VLMs are not specifically tailored to human behavior, limiting their effectiveness in capturing critical human-centric details essential for safe driving.

In this work, we aim to answer three core questions regarding human behavior understanding tasks in autonomous driving scenarios: \ding{182} What aspects of human behavior are critical 
to autonomous driving? \ding{183} How effectively do current approaches model human behaviors in autonomous driving contexts? \ding{184} How can a comprehensive benchmark advance the development of human behavior understanding algorithms?
To this end, we propose \bfours, a large-scale unified benchmark explicitly designed for comprehensively understanding various human behaviors in driving scenarios. 
\ours~includes rich annotations generated by a human-in-the-loop annotation pipeline, enabling scalable and precise labeling from diverse data sources using only monocular video inputs.
Specifically, we have collected 1.73M frames featuring 57K human instances from source videos obtained from Waymo~\cite{sun2020scalability}, YouTube, and self-collected data. 
The dataset provides detailed annotations covering: \textbf{(1)} human motion and trajectory; \textbf{(2)} text descriptions of human motions generated using templates and VLMs;  \textbf{(3)} critical human behaviors 
extracted via VLMs, along with question-answer (QA) pairs designed to benchmark driving-oriented and generalist VLMs.

Our contributions can be summarized as follows:
\begin{itemize}[leftmargin=2em, itemsep=-1pt]
\vspace{-3pt}
    \item We introduce \textbf{\ours}, a unified, human-centric dataset that provides a comprehensive understanding of humans' behaviors in driving scenarios that can be used as a benchmark for a range of human-centric understanding tasks.
    \item We develop a scalable, human-in-the-loop annotation pipeline employing multi-source fitting strategies to produce accurate labeling across diverse video sources, ranging from driving videos and general YouTube videos to self-collected streams.
    \item We evaluated baseline methods of human behavior understanding and analyze their performance, we further demonstrated that our dataset helps these methods achieve better performance.
\end{itemize}

\section{Related Works}
\vspace{-3mm}
\subsection{Human Motion}
\vspace{-2mm}

Human motion is essential to autonomous driving. We categorize human motion representations into 2D and 3D representations. 2D human motion~\cite{jiang20242d, belagiannis2017recurrent, luo2021rethinking, li2021tokenpose, jin2020whole} leverages keypoints or heatmaps to mark the local body motion on the image. 
For 3D representations, the SMPL series~\cite{SMPL:2015, MANO:SIGGRAPHASIA:2017, SMPL-X:2019} provide compact and expressive representation via learned parameters. 
While most human motion datasets focus on general human motions~\cite{ionescu2013human3, von2018recovering, xu2024motionbank, mahmood2019amass, lin2023motion}, there are several datasets are specially designed for driving scenarios~\cite{wangpedestrian, liu2024learning, kim2024text}. However, these datasets mainly focus on the human movements and their text description, the behaviors of humans remains unavailable.
Based on the representation and datasets, several efforts have been put into human motion reconstructions from temporally aligned multi-view cameras~\cite{huang2021dynamic}, unaligned multi-view cameras~\cite{dong2020motion}, and monocular cameras~\cite{luvizon2023scene, li2022d, ye2023decoupling, yuan2022glamr}. Other works have explored human motion generation from action labels~\cite{cervantes2022implicit} or text~\cite{tevet2022human, zhang2024motiondiffuse}. However, due to the lack of high-quality data, there is little work that specifically generates human motion in driving scenarios.

\vspace{-2mm}
\subsection{Human Behavior Understanding}
\vspace{-3mm}
Understanding human behavior in driving situations is essential for driving safety. Although there are some datasets and methods to understand human behavior and actions~\cite{zhang2024incorporating, rai2021home, punnakkal2021babel, wang2012mining, li2010action, soomro2012ucf101, niebles2010modeling, kay2017kinetics, marszalek2009actions}, they mainly focus on recognizing human actions in general or sports scenes. While sharing some common behavior that concerns driving safety, they mainly focus on general actions like shaking hands, dancing, or running. The behaviors specifically concerning driving safety remain unexplored. In autonomous driving, besides motion reconstruction, there are some approaches and datasets for understanding several aspect of human behaviors such as (1) human trajectory prediction~\cite{zhang2024incorporating, goncalves2022auxformer, medina2024context, wang2024gcnext, guo2023back, mao2020history}, where models are required to predict the future trajectory from previous ones, or (2) human intention prediction~\cite{zhang2023trep, osman2023tamformer, sharma2023visual, rasouli2019pie, kotseruba2021benchmark, Rasouli2017IV}, where pedestrians are simply classified into two states - crossing the street and not crossing the street. Besides the binary classification of crossing the street, there are several datasets~\cite {kwak2017pedestrian, quintero2015pedestrian, schneider2013pedestrian, rasouli2017they} that provide more detailed behavior labels such as stopping, glancing, or running. However, these works still focus on specified aspects of human behavior, such as the posture and action when a pedestrian is crossing the street. Recently, the development of vision language models (VLMs)~\cite{touvron2023llama,lin2023vila, chen2024longvila, liu2024nvila, liu2023llava, liu2024llavanext, liu2023improvedllava, zhang2024videoinstructiontuningsynthetic, bai2023qwen} enables question-answering based on images or videos, making human behavior understanding more flexible. There have been many specialists driving VLMs~\cite{ma2024dolphins, sima2024drivelm, chen2023tem,shao2024lmdrive, wang2024omnidrive, yuan2024rag, chen2024driving} and autonomous driving QA datasets~\cite{marcu2024lingoqa, arai2024covla, nie2024reason2drive, chen2024driving, inoue2024nuscenes, qian2023nuscenes}. However, these models and the datasets are designed for general VQA tasks for autonomous driving. The comprehensive understanding of human behaviors remains unexplored. We show a comparison of related datasets in Tab.~\ref{table:dataset_comp}.

\vspace{-2mm}
\subsection{Autonomous Driving Datasets}
\vspace{-3mm}
Autonomous driving has been one of the most popular research topics in recent years. There are several datasets that are specially created for developing and evaluating autonomous driving algorithms~\cite{geiger2013vision, sun2020scalability, caesar2020nuscenes, maddern20171}. These datasets are typically collected from a vehicle mounted with multiple sensors, such as multi-view cameras, LiDARs, RaDARs, IMU, etc., supporting autonomous driving tasks such as 2D and 3D object detection, semantic segmentation, depth estimation, and planning. Recently, several works have focused on some specific scenes in autonomous driving, such as the accident~\cite{fang2021dada}, snowy scenes~\cite{chen2023desnow}, and foggy scenes~\cite{sakaridis2018semantic}.
Some datasets have provided the labeling of several aspects of human behaviors in driving scenarios, such as human motion and trajectory~\cite{kim2024text, wangpedestrian, liu2024learning}, intention of crossing the street~\cite{rasouli2017they, bhattacharyya2021euro, rasouli2019pie}, or in the forms of general VQA~\cite{arai2024covla, qian2023nuscenes, inoue2024nuscenes, sima2024drivelm}. However, these datasets only investigate some specific human behaviors, and the comprehensive understanding of human behavior remains unexplored.

\section{The \ours~Dataset}
\label{sec:dataset}

\begin{figure*}[ht]
    \centering
 \includegraphics[width=0.95\linewidth]{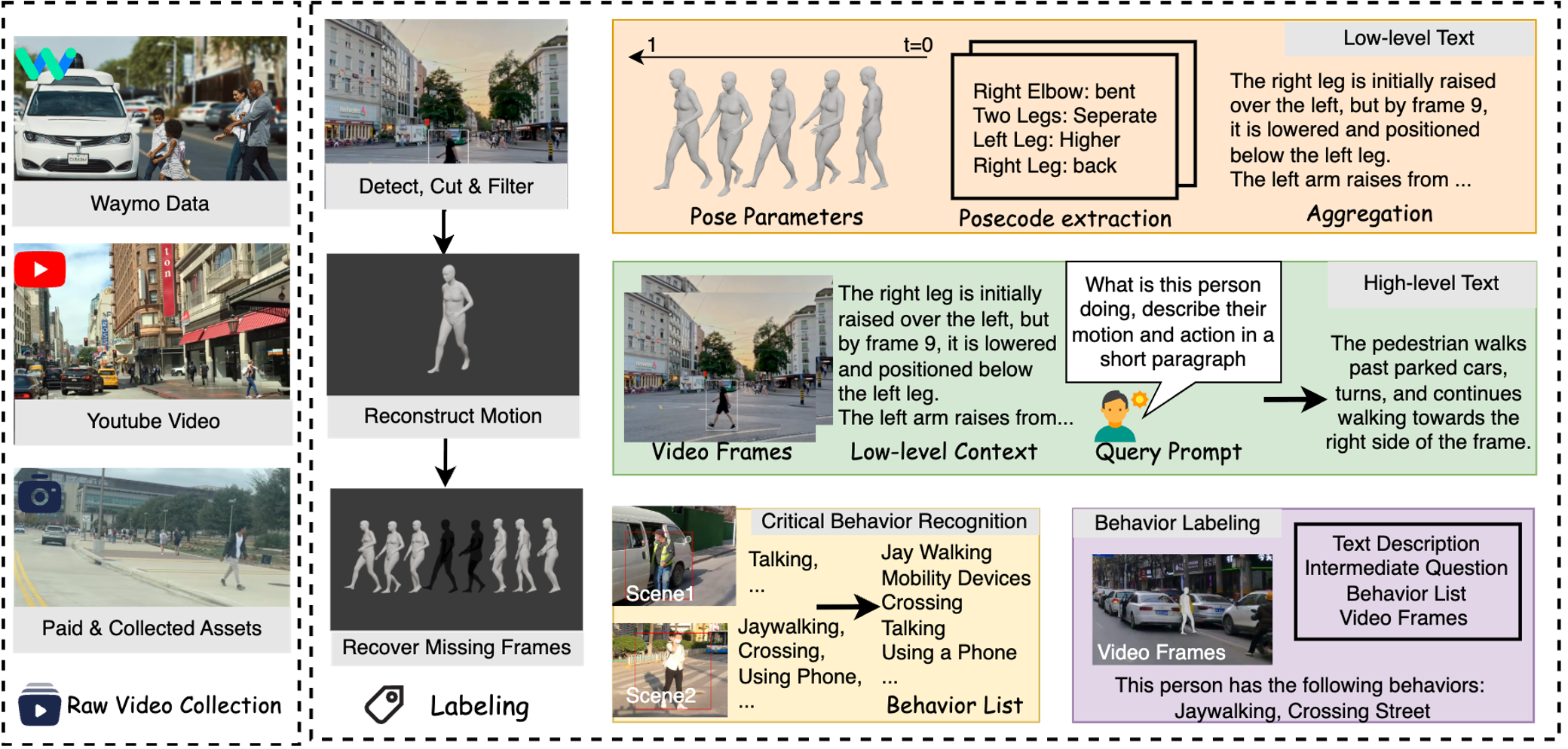}
 \vspace{-2mm}
    \caption{\textbf{Data Collection and Annotation.} (Left) We collect data from three sources: the Waymo dataset, the YouTube videos, and the self-collected or paid driving videos. (Right) We demonstrate the annotation pipeline; we first filter and cut the raw videos based on the rough human detection results. Then we reconstruct the SMPL motion for each detected frame. The missing frames are further recovered by an interpolation procedure. For the labeling of text descriptions, we leverage low-level text as a bridge between the SMPL parameters and the semantic label. Then we generate the high-level text from the low-level ones. We recognize the critical behavior lists leveraging a VLM, based on the visual and text information. We label the behaviors for each human instance using a VLM that is fine-tuned on a small human-labeled subset.
}
    \label{fig:pipeline}
    \vspace{-3mm}
\end{figure*}

\begin{figure*}[!ht]
    \centering
 \includegraphics[width=\linewidth]{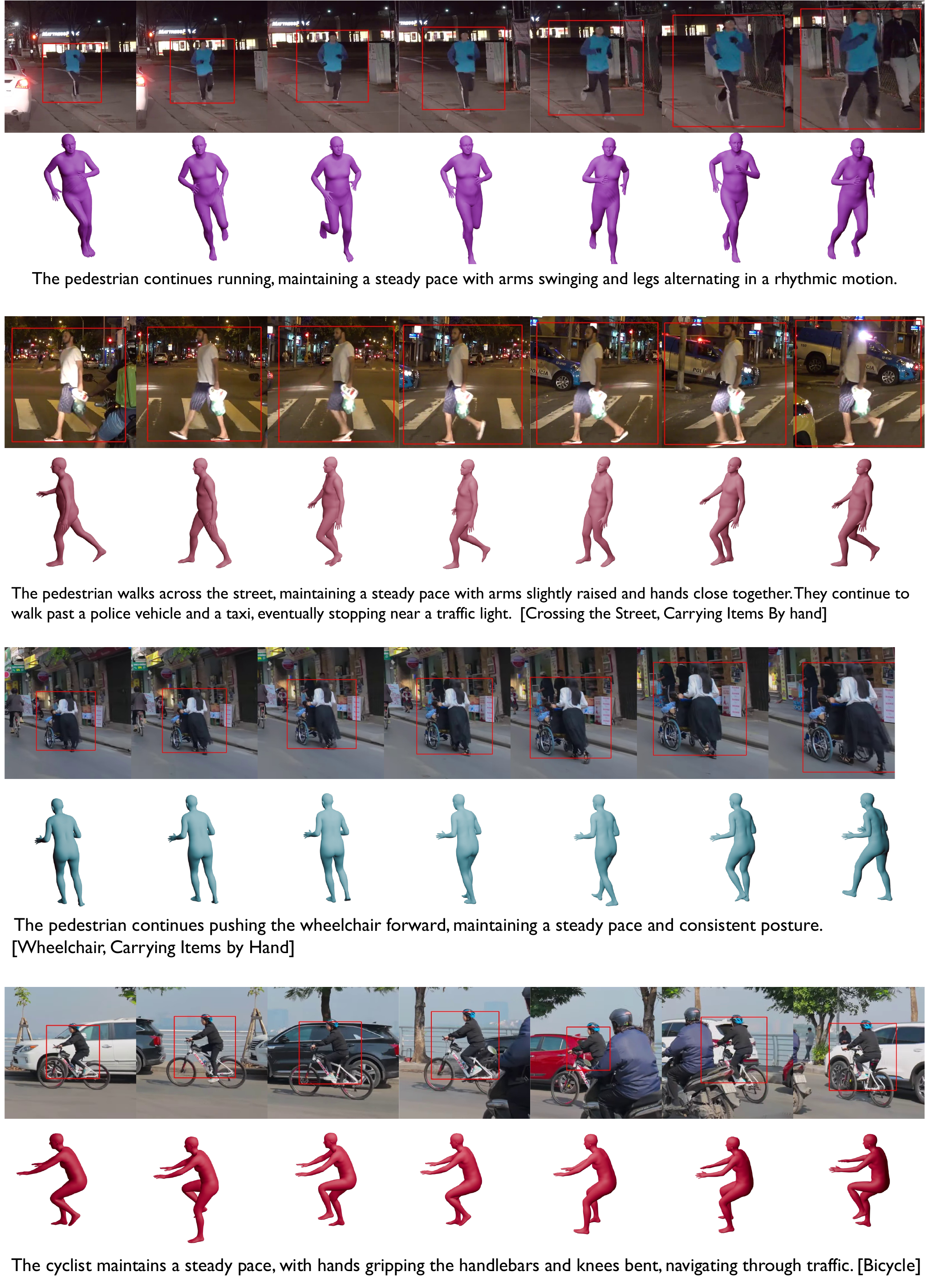}
    \caption{\textbf{Visualization of the \ours~dataset.} 
    For each human instance, the first line shows the video frames that is sampled from the video clips. The human instance is highlighted using a red bounding box. We crop and zoom-in the human instance for a clearer view. Under each of the frames shows the corresponding human motion rendered as mesh, followed by the text description for the human motion and the behavior labels.
    }
    \label{fig:supp-vis-data}
\vspace{-3mm}
\end{figure*}

\vspace{-3mm}
\paragraph{Overview.} As illustrated in Fig.~\ref{fig:supp-vis-data}, we propose \bfours, a comprehensive human-centric benchmark with rich annotations, emphasizing the criticalness of human behavior understanding in autonomous driving. 
We built our dataset using high-quality videos from various sources. 
Then we applied a scalable annotation pipeline that only involves minor human effort to get the rich annotations from the collected videos. The annotation for each video clip includes the 3D motion with trajectory, 
intention, high- and low-level text description, and critical behavior labels concerning driving safety. We will present our data collection details in Section~\ref{sec: data_collection}. We then introduce the annotation pipelines and data analysis of human motion and trajectory (Sec.~\ref{sec:motion}), text description (Sec.~\ref{sec:text}), and intention and critical behaviors (Sec.~\ref{sec:intention}). We show the statistics of the dataset in Sec.~\ref{sec:stat}.

\subsection{Data Collection}
\label{sec: data_collection}

\paragraph{Videos Acquisition}
We collected raw videos of 1.73M frames in total from three kinds of data sources as follows:
\underline{Autonomous Driving Data} contains multi-modality sensor information and rich annotation for generic autonomous driving tasks such as object detection, depth estimation, etc. We collected \xtodo{1.7 hours} videos from Waymo~\cite{sun2020scalability}, which consist of \xtodo{73K} frames. 
\underline{In-the-wild Data} includes first-person driving videos that are publicly available on the Internet. We collected 10 hours of YouTube videos, each with a CC license, consisting of \xtodo{318.25K} frames with a resolution ranging from \xtodo{1080p} to \xtodo{2k}. 
\underline{Self-collected Data} is driving recordings collected by ourselves or from paid sources. We collected \xtodo{66.5} hours of videos, consisting of \xtodo{2393.96 k} frames and the resolution varies from \xtodo{1080p} to \xtodo{4k}.

\paragraph{Video Cutting and Filtering}
Directly applying the annotation pipeline to the entire video can be expensive. We first roughly detect the human presence and filter out the frames that lack human presence. Specifically, we apply a human detector on the raw video at 1 FPS, then we separate the raw video into fragments separated by the no-human-presented frames. We then filter out the fragments that are less than 10 seconds.
\vspace{-2mm}
\subsection{Motion and Trajectory}
\label{sec:motion}
\vspace{-3mm}
\paragraph{Motion Extraction.} 
Human motion provides rich information about human actions and behaviors. 
We extract motion sequences from the raw videos collected as in Sec.~\ref{sec: data_collection} and recover human trajectories from the motion sequences. For human motion extraction, we reconstruct the human motions parameterized by the SMPL~\cite{SMPL:2015} representation. 
Formally, given a person in $T$ frames, their motion is described as a SMPL parameter sequence {$\mathcal{S}=\{S_t\in\real^{n\times m}|t\in\{1,...,T\}\}$}, $S_k$ defines the motion at time $t$. We follow the detect-and-reconstruct schema for motion reconstruction from video. We first leverage an object detector to detect humans. Then we track each individual during the detected frames. We finally reconstruct their SMPL parameters following Wham~\cite{shin2024wham}. We employ different procedures to track the individuals, considering different data sources. For sources that provide bounding boxes of humans, e.g., the autonomous driving datasets, we leverage the bounding boxes as a prior. We compare the provided bounding boxes with the detected ones, select those that have at least \xtodo{0.2} IoU overlap with one of the detected bounding boxes as reconstruction candidates. Then we delete the frames for each individual whose bounding box intersects others with IoU higher than \xtodo{0.2}, which indicates some occlusion.
For the in-the-wild and self-collected sources, we unify the videos to 10 FPS and extracted the 2d keypoints of each individual and fused them with the detected bounding boxes as the tracking results.

\begin{figure*}[ht]
    \centering
     \includegraphics[width=0.99\linewidth]{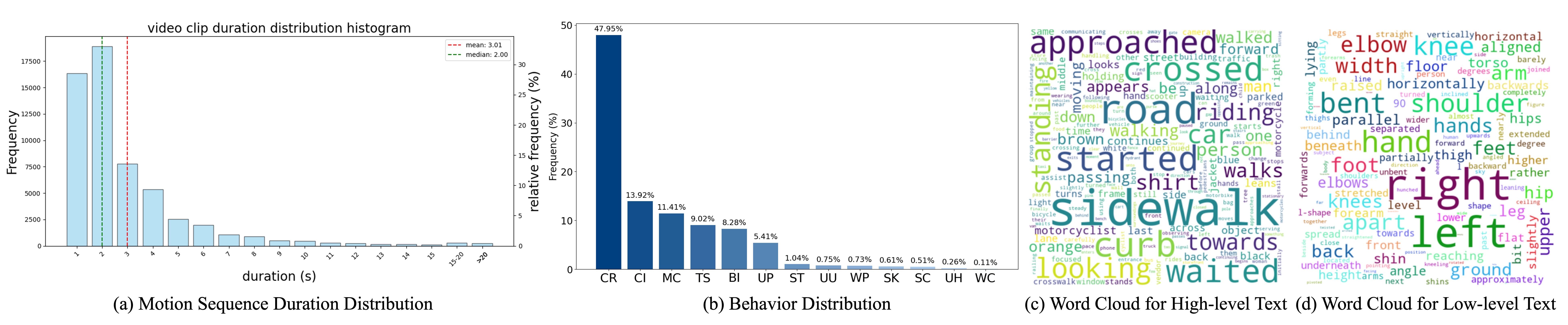}
    \caption{\textbf{Statics of \ours.} The average duration of motion sequences is 3s. The most common behavior is crossing the street, while the rarest behavior is using a wheelchair. Behavior definition, please refer to Sec.~\ref{sec:intention}.}
    \label{fig:statics}
\end{figure*}

\begin{table*}
\centering
\caption{\textbf{Comparison of Related Datasets.} We compare our dataset with a related dataset of general motion or driving scenes. From left to right, the columns are the dataset name, total frame count or time duration, human count, providing labels of motion, trajectory, VQA pairs, and text descriptions, and the number of behavior classes. $^\ddagger$ represents a general motion dataset. Dataset labeled with $^*$ in the ``Instance" column captures the motion from real participants. $^\dagger$ means upper-bound. Datasets labeled with ``Unstructured" in the last column do not provide explicit human behavior labels but involve them in the QA pairs or captions. Our dataset supports all four tasks and provides the most behavior classes among all of the datasets.}
\label{table:dataset_comp}
\setlength{\tabcolsep}{3pt}
\resizebox{\textwidth}{!}{
\begin{tabular}{lcccccccc}
    \toprule
    Dataset & Frames / Duration (s) & Instances & Motion & Trajectory& VQA & Text  & Behaviors \\
    \midrule
    PIE~\cite{rasouli2019pie}& 293k / - & 1.3k & \xmark & \xmark  & \xmark & \xmark & 1  \\
    Euro-PVI~\cite{bhattacharyya2021euro}& 83k / - & 7.7k & \xmark & \xmark & \xmark & \xmark & 1  \\
    PMR~\cite{wangpedestrian}& 225k / - & 54$^*$ & \checkmark & \checkmark & \xmark & \xmark  & \xmark \\
    CityWalker~\cite{liu2024learning}& - / 110k & 120k & \checkmark & \checkmark& \xmark & \checkmark & \xmark \\
    BlindWays~\cite{kim2024text}& 300k / 10k & 11$^*$ & \checkmark & \checkmark & \xmark & \checkmark & Unstructured \\
    3DPW$^\ddagger$~\cite{vonMarcard2018} & 51k / 1.7k & 7$^*$ & \checkmark & \checkmark & \xmark & \xmark & \xmark \\
    Human3.6M$^\ddagger$~\cite{ionescu2013human3}& 3.6M / - & 11$^*$ & \checkmark & \checkmark & \xmark & \xmark & 15$^\ddagger$ \\
    JAAD~\cite{rasouli2017they}& 82k / 33k$^\dagger$ & 2.2k & \xmark & \xmark & \xmark & \xmark & 11  \\
    Drama~\cite{malla2023drama}& - / 36k & - & \xmark & \xmark & General Driving & \checkmark & Unstructured  \\
    CoVLA~\cite{arai2024covla}& 6M / -& -   &\xmark &  \xmark & General Driving & \checkmark & Unstructured \\
    DriveLM-nuScenes~\cite{sima2024drivelm}& 4.8k / - & - & \xmark & \xmark & General Driving & \checkmark & Unstructured \\
    nuScenes-QA~\cite{qian2023nuscenes}& 340k / - & - & \xmark & \xmark & General Driving & \checkmark & \xmark  \\
    
    \ours~(Ours) & 1.73M / 173K & 57K&  \checkmark & \checkmark & Human-Centric & \checkmark &13& \\
   
  \bottomrule
\end{tabular}}
\vspace{-2mm}

\end{table*}
\vspace{-2mm}
\paragraph{Trajectory from Motion.} The trajectory of the human is represented as the a sequence of point sets $\{P_1, ..,P_p, ...P_F\}$, each point set contains $n$ points representing the 3D location of human parts $P_t=\{p_k\in\real^3 |k\in\{1,...n\}\}$. The sequence preceding $P_p$, denoted as $S_p$, represents past motion, while the sequence following $P_p$ is referred to as $S_f$, representing future motion. And $S_f$ is initially padded with zeros. A trained model $F_{pred}$ predicts $S_f$ given the input $S_p$. We extract the trajectories of each human from their global motion sequence. 
\vspace{-2mm}
\paragraph{Motion Completion.}
The motion sequence of a person can lack a few frames in the middle due to occlusion. We present a missing motion prediction procedure to complete the missing frames. Specifically, given reconstructed SMPL parameter sequence $\mathcal{S}=\{S_1, S_2,...,S_n\}$, and $m$ missing frames $k, k+1,...,k+m$, the predicted SMPL parameter ${\hat S_{k},...\hat S_{k+m}}$ is as follows:
$$
\hat S_{k+j} =\frac{\sin((1-\frac{j}{m})\theta)}{\sin(\theta)}\hat S_{k-1} + \frac{\sin(\frac{j}{m}\theta)}{\sin(\theta)}\hat S_{k+m+1}
,~~~
\cos(\theta) = S_{k-1} \cdot S_{k+m+1}
$$

\subsection{Hierarchical Text Annotation}
\label{sec:text}
\vspace{-2mm}
In addition to the parameterized SMPL motion, the semantic understanding of human behaviors is also critical. We use text as a semantic level description of the human motion. To narrow the gap between the semantic description and the SMPL parameter, we employ a hierarchical text annotation approach. We first convert the SMPL parameters to an element-level text description for each part of the body at each frame in a rule-based schema. Then we utilize large language models to aggregate the low-level description of the person over time. Based on the detailed low-level motion description, combined with the video clip, we abstract the high-level descriptions for each motion sequence. The details of hierarchical text annotation are introduced in the supplementary materials.
\underline{Low-level Text Annotation:} As shown in~\ref{fig:pipeline}, the low-level text describes the movement of each body part in detail. Following~\cite{delmas2022posescript}, we generate the low-level text description for the SMPL motion $S_t$ at each frame $t$ by calculating the angle, distance, position relation, etc. of different body parts. We then aggregate the movement of each body part over time to get the low-level description of human motion. 
\underline{High-level Text Annotation:} High-level captions provide a semantic-level description of human action and motion. We generate one high-level description for the motion sequence of each person leveraging large vision language models. Specifically, we provide the VLM eight frames uniformly sampled over time, including the images and the low-level text descriptions.

\vspace{-2mm}
\subsection{Critical Behaviors}
\label{sec:intention}
\vspace{-2mm}
Understanding human behaviors --- for example, crossing the street --- is critical for autonomous driving algorithms. We model the critical behaviors as binary attributes, indicating whether a person is subject to the corresponding behavior. 
\vspace{-2mm}
\paragraph{Critical Behavior Recognition} Before we can assign the values to each attribute, we should first answer the question: what behaviors are critical to autonomous driving? One of them might be whether a person is going to cross the street, which is known as the intention prediction task in autonomous driving. {As illustrated in Fig.~\ref{fig:pipeline}}, we recognize the critical behaviors by leveraging VLMs. Specifically, we sample $n$ video clips from the dataset. For each video clip, we ask the VLM to recognize the critical behaviors in the scene. Lastly, the answers are collected, and a VLM is instructed to summarize the critical behaviors for autonomous driving. We recognize 13 behaviors, namely walking pets (WP), talking (TS), using a phone (UP), using an umbrella (UU), using headphones (UH), carrying items in hand (CI), crossing the street (CR), using wheelchair (WC), using stroller (ST), riding bike (BI), riding scooter (SC), using skateboard (SK), and riding motorcycle (MC).

\vspace{-2mm}
\paragraph{Critical Behavior Labling}
\begin{wraptable}[9]{c}
{0.5\textwidth}
\centering
\vspace{-6mm}
\caption{
\small
\textbf{Motion Generation Evaluation.} Given the high FID distance, the generic text-to-motion models cannot properly generate motions in driving scenes.}
\label{tab:motion_gen}
\small
  \resizebox{0.48\textwidth}{!}{
\begin{tabular}{ccc}
\toprule
Model & FID $\downarrow$ & Multi Modality \\
\midrule
Real & 0.002 & - \\
MotionDiffuse~\cite{zhang2024motiondiffuse} &  39.275 & 2.36 \\
MotionGPT~\cite{jiang2023motiongpt} & 27.059 & 5.42   \\      
\bottomrule
\end{tabular}
}
\end{wraptable}
Given the recognized behavior set $\Set B = \{\Mat b_k|k\in\{1,...,m\} \}$, the label of critical behaviors for a person $h$ is defined as the subset $\Set B_h\subseteq \Set B$ in which the behaviors hold for the person. For each instance, we enumerate each element in the behavior set $b_k\in\Set B$ and construct a corresponding question $q_k$. Then we provide the corresponding frames to a VLM and ask it about the question $q_k$. Based on the answer, we append $b_k$ to the behavior set $\Set B_h$ for the person. Directly applying pre-trained VLMs can suffer from noisy labeling results. To alleviate this, we perform a human-in-the-loop labeling strategy. We randomly selected 10\% of the dataset and employed human annotators to label whether the given person has certain behaviors. Then we use the human-labeled data to fine-tune the annotation VLM, which will further be applied to the rest of the unlabeled data. The details of the human annotation are described in the supplementary materials.

\begin{table*}[]
\centering
\small
\caption{\textbf{Evaluation of Human Behavior Visual QA.} We construct close-ended questions where the model is asked to select whether the person subjects to certain behaviors. We report the F1 score for each behavior, and then an instance-averaged F1 score is used to evaluate the overall performance. Behavior definition please refer to Sec.~\ref{sec:intention}.}
\label{table:behavior}
\resizebox{\textwidth}{!}{
\begin{tabular}{lccccccc|cccccc|c}
\toprule
Baseline & WP & TS & UP & UU & UH & CI & CR & WC & ST & BI & SC & SK & MC & Micro-F1 \\
\midrule
Phi-4-multimodal & 42.9 & 35.6 & 24.6 & 90.9 & 15.4 & 39.4 & 26.1 & 100.0 & 31.2 & 56.3 & 66.7 & 33.3 & 77.1 & 45.5 \\
MiniCPM-o-2\_6 & 75.0 & 27.4 & 58.6 & 100.0 & 44.4 & 33.4 & 26.4 & 85.7 & 77.8 & 88.9 & 80.0 & 50.0 & 73.6 & 52.2 \\
Dolphins & 42.9 & 2.3 & 1.0 & 22.2 & 40.0 & 0.3 & 16.9 & 40.0 & 6.5 & 1.7 & 40.0 & 57.1 & 1.3 & 3.1 \\
Qwen2-VL-7B & 66.7 & 2.8 & 63.3 & 100.0 & 25.0 & 30.7 & 66.7 & 100.0 & 76.9 & 69.3 & 80.0 & 66.7 & 80.7 & 52.1\\
Qwen2.5-VL-7B & 42.9 & 16.4 & 36.4 & 76.9 & 15.4 & 40.6 & 32.4 & 85.7 & 34.3 & 47.9 & 66.7 & 25.0 & 73.9 & 44.7 \\
InternVL2-8B & 33.3 & 1.5 & 17.2 & 100.0 & 15.4 & 11.2 & 6.3 & 85.7 & 25.0 & 54.7 & 66.7 & 28.6 & 68.8 & 27.7 \\
InternVL2.5-8B & 75.0 & 14.5 & 32.6 & 100.0 & 15.4 & 27.3 & 24.4 & 100.0 & 76.9 & 72.5 & 80.0 & 50.0 & 75.5 & 42.2 \\
Mantis-8B-SigLIP & 80.0 & 28.1 & 68.6 & 100.0 & 28.6 & 52.5 & 22.6 & 100.0 & 93.3 & 91.0 & 80.0 & 50.0 & 87.0 & 58.4 \\
Aya-Vision-8B & 46.2 & 23.8 & 43.3 & 100.0 & 40.0 & 40.4 & 21.3 & 100.0 & 75.0 & 77.6 & 80.0 & 50.0 & 29.1 & 38.6 \\
Idefics3-8B-Llama3 & 42.9 & 18.5 & 19.7 & 90.9 & 15.4 & 23.5 & 24.4 & 85.7 & 33.3 & 64.1 & 80.0 & 28.6 & 84.6 & 40.1 \\
Pixtral-12b & 36.4 & 22.9 & 18.4 & 90.9 & 18.2 & 25.6 & 21.7 & 100.0 & 37.0 & 58.5 & 66.7 & 33.3 & 78.7 & 41.1 \\
Gemma-3-12B-it & 53.3 & 36.6 & 32.2 & 90.9 & 25.0 & 51.6 & 12.1 & 100.0 & 41.0 & 63.3 & 66.7 & 40.0 & 87.1 & 52.9 \\
Deepseek-vl2-small & 42.9 & 19.9 & 54.1 & 76.9 & 15.4 & 30.6 & 37.6 & 85.7 & 34.3 & 37.5 & 66.7 & 28.6 & 44.0 & 34.9 \\
Kimi-VL-A3B & 46.2 & 21.2 & 39.4 & 76.9 & 20.0 & 35.8 & 23.0 & 85.7 & 46.7 & 66.7 & 66.7 & 33.3 & 85.7 & 47.7 \\
GPT4o-mini & 85.7 & 67.7 & 62.5 & 100.0 & 40.0 & 55.0 & 58.4 & 100.0 & 66.7 &  90.1 & 80.0  &  50.0 & 65.36 & 64.8\\
\bottomrule
\end{tabular}}
\end{table*}

\begin{table}[!htb]
    \begin{minipage}{.65\linewidth}
      \caption{\textbf{Evaluation of Motion Prediction Baselines.} We evaluate the motion prediction baselines on \ours-T. We leverage the pre-trained weights and evaluate them on our dataset without fine-tuning. All of the baselines generate plausible trajectory predictions, and PhysMoP achieves the best performance.}
      \label{tab:motion_prediction}
      \small
      \centering
      \resizebox{\textwidth}{!}{
        \begin{tabular}{cccccccccc }
    \toprule
     &  &  &  & MPJPE$\downarrow$ &  &  &  &  \\
    frame\_id&1 & 3 & 7 & 9 & 13 & 17 & 21 & 24\\
    \midrule
    PhysMoP~\cite{zhang2024incorporating}  & 0.4 & 1.7 & 9.0 & 14.4 & 26.3 & 36.2 & 45.3 & 54.3\\
    AuxFormer~\cite{xu2023auxiliary}& 17.0 & 32.7 & 47.8 & 60.0 & 71.1 & 79.0 & 84.3 & 86.1\\
    CIST-GCN~\cite{medina2024context}& 18.5 & 25.3 & 37.2 & 40.8 & 46.2 & 46.6 & 46.8 & 47.4\\
    
  \bottomrule
\end{tabular}
}
    \end{minipage}%
    \hfill
    \begin{minipage}{.3\linewidth}
      \centering
        \caption{\textbf{Benefiting Behavior VQA.} By finetuning on \ours, the baseline model (QWen2.5-VL) shows a significant improvement on averaged accuracy and F1-score.}
        \label{tab:bene_vqa}
      \resizebox{\textwidth}{!}{
        \begin{tabular}{ccc}
\toprule
Model                         &       Accuracy$\uparrow$                      & F1-Score $\uparrow$                    \\
\midrule
Baseline                            & 35.31                          &        44.72            \\
Finetuned                            & 67.77                          &        68.54            \\      
\bottomrule

\end{tabular}
}
    \end{minipage} 
\end{table}

\vspace{-2mm}
\subsection{Statistics}
\label{sec:stat}
\vspace{-3mm}

As shown in Fig.~\ref{fig:statics} (a) and Tab.~\ref{table:dataset_comp} (last row), our dataset provides \xtodo{48 hours} human motion sequences and corresponding video clips in total. The frame rate of both the motion sequence and video clip is \xtodo{10 Hz}, thus the total frames sum up to \xtodo{1.73M}. The durations of motion sequences range from \xtodo{1} to \xtodo{12} seconds. The mean and median durations are \xtodo{3.01} and \xtodo{2} seconds. Regarding the text descriptions, the average length of low- and high-level descriptions is \xtodo{15} and \xtodo{33} words, respectively. The world clouds for the two kinds of descriptions are illustrated in Fig.~\ref{fig:statics} (c) and (d). Considering human behaviors, which is one of the characteristics that distinguishes \ours~ from previous ones, the most frequent behavior is crossing the street, followed by carrying items, and the least frequent one is using a wheelchair, as illustrated in Fig.~\ref{fig:statics} (b).

\vspace{-3mm}
\section{Tasks}
\label{sec:tasks}
\vspace{-3mm}
\ours~ supports multiple human-centric tasks.
In this section, we will present the definition of tasks and corresponding evaluation metrics in our experiments.

\vspace{-3mm}
\paragraph{Motion Prediction.}
Understanding the historical motion and predicting future ones is essential for the safety of autonomous driving. Following PhysMoP~\cite{zhang2024incorporating}, we leverage the sequence of human motion keypoints as the representation. 
Specifically, human motion involves $n$ frames $M_1,...M_n$, each of which represents the global location of human joints at frame $I_i$. Each motion frame $M_i$ consists the location of $m$ key joint point on human body, 
i.e. $M_i=\{p_k\in\|k\in\{1,...,m\}\}$. The human motion prediction task is about predicting future motion frames from historical ones, i.e., predicting $M_{t1+1},...M_{t_1+t2}$ from $M_0,..M_{t1}$. We employed two widely used metrics to evaluate the baseline methods: (1) Mean Per Joint Position Error (MPJPE), which measures the mean 3D Euclidean distance between the predicted and ground truth joint positions after aligning the root joint; (2) and the ACCL metric, measuring the acceleration error averaged over time to measure the physical plausibility of the predicted motion.

\vspace{-3mm}
\paragraph{Motion Generation.}
Though motion is an informative and compact representation of human actions, collecting them in the real driving scene is expensive and even dangerous for some behaviors. Motion-from-text generation can be a very efficient and effective way to augment motion data. Specifically, given a text description of motion, the algorithm is supposed to generate motion sequences that are subject to the text description. We thus test the capacity of existing current text-to-motion approaches to generate human motions in a driving scene. We follow MotionDiffuse~\cite{zhang2024motiondiffuse} to model motion sequences by converting from the SMPL parameters. We leverage the high-level description as the text prompt to generate the motions. We employ FID~\cite{heusel2017gans} and multi-modality as evaluation criteria. The former evaluates the distributional distance between the generated motions and the ground-truth motions, while the latter measures joint position differences among 32 motion sequences generated from the same text description.

\vspace{-3mm}
\paragraph{Behavior VQA.}
Humans in the street can have multiple characteristics and behaviors. Unlike previous related tasks, such as intention prediction, where the behavior is simply classified into binary labels --- crossing the street or not crossing the street, we propose a new behavior set that provides more comprehensive aspects of human behaviors. To make the tasks flexible and easy to extend, we formulate the task as a visual question-answering (VQA) task. We label humans with 13 behaviors, which are all binary labels, as mentioned in Sec.~\ref{sec:intention}. We construct close-ended questions for all labeled behaviors using a template. An example is ``Is the person in the video riding a bike? (A) Yes (B) No." We then evaluated the accuracy of the baseline methods. We employed the accuracy and F1-score as evaluation metrics.

\vspace{-2mm}
\section{Experiments}
\label{sec:exp}
\vspace{-2mm}
In this section, we evaluate recent methods for human behavior understanding in our dataset. Specifically, we evaluated methods related to human motion prediction, text-to-motion generation, and human behavior VQAs. We analyze their performance on their corresponding tasks.
\vspace{-2mm}
\subsection{Dataset Splitting}
\vspace{-2mm}
\ours~is split into three subsets, namely \ours-V, \ours-H, and \ours-T, each consisting of \xtodo{47k}, \xtodo{9.5k}, and \xtodo{840} human instances, representing the VLM-labeled, human-labeled, and testing data.
We sample the human-labeling subset (which should be \ours-H + \ours-T) from three strategies. (1) in-video-clip sampling:
For some video clips, we randomly sample some individuals from each of them as human-labeling data and the rest as \ours-V data.
(2) in-video sampling: we sample some video clips from each raw video and use all the individuals as human-labeling data, and the rest video clips serve as \ours-V data. (3) out-of-video sampling: we randomly choose some raw videos, and the entire videos serve as the human-labeling dataset. Similarly, we split the \ours-H and \ours-T subsets from the human-labeling data.
In the experiments, without specifically mentioning, we use both \ours-V and \ours-H as training data, denoted as \ours.
\vspace{-2mm}
\subsection{Baselines}
\vspace{-2mm}

\underline{Motion Prediction:}
For motion prediction, we employ PhysMoP~\cite{zhang2024incorporating}, CIST-GCN~\cite{medina2024context}, and AuxFormer~\cite{goncalves2022auxformer} as the baselines. Following the settings in~\cite{zhang2024incorporating}, we randomly select 50 continuous frames from each motion sequence, using the first 25 frames as the input for the baselines and comparing their output with the remaining 25 frames.  
\underline{Motion Generation:}
We choose MotionDiffuse~\cite{zhang2024motiondiffuse} and motionGPT~\cite{jiang2023motiongpt} as our baseline. For both methods, we provided the high-level descriptions from our dataset as the text prompt. Following motionGPT~\cite{jiang2023motiongpt}, we only use motion sequences that are within 20 to 196 frames.
\underline{Behavior VQA:} For the 13 critical behaviors, there is no specialist model that is trained to predict all of them. Thus, we employed generalist vision-language models as our baselines.
We provided the vision language models 4 to 6 frames sampled from the corresponding video and a close-ended question constructed from the behavior labels as input,
and evaluated whether the VLMs can accurately recognize these essential behaviors and answer the provided question. We employ multi-image querying VLMs, including Phi-4-multimodal~\cite{abouelenin2025phi}, MiniCPM~\cite{yao2024minicpm}, Qwen~\cite{wang2024qwen2, bai2025qwen2}, InternVL~\cite{chen2024expanding}, Mantis~\cite{jiang2024mantis}, Idefics3~\cite{laurençon2024building}, Pixtral~\cite{agrawal2024pixtral}, Gemma3~\cite{team2025gemma3}, Deepseek-VL2~\cite{wu2024deepseekvl2}, Kimi-VL~\cite{team2025kimi}.

\vspace{-2mm}
\subsection{Baseline Evaluation}
\vspace{-2mm}

\underline{Motion Prediction:}
We evaluated the pre-trained baselines on \ours-T, to unify the time argument used in computing the MPJPE metric, we use the frame id to replace it. All baselines are evaluated on the same frames.
The results are shown in Tab.~\ref{tab:motion_prediction}.
All of the approaches generate plausible results, even though they are not specially trained for driving scenes.
Among them, PhysMoP~\cite{zhang2024incorporating} achieves the best performance in our dataset.
\underline{Motion Generation:}
As shown in Tab.~\ref{tab:motion_gen} and Fig.~\ref{fig:motion_generation_comparsion} (up row), motion generation models pre-trained on generic human motion datasets are not capable of generating plausible motions in driving scenes, due to the domain gap between the general motions and motions in the street. We further provide visualization results of these baselines in the supplementary material.
\underline{Behavior VQA:} 
We evaluated the ability of mainstream vision-language models to recognize human behaviors in driving scenes. We constructed closed-ended questions based on the behaviors and provided 4 to 6 frames to the VLMs. We evaluated the VLMs based on their correctness in answering these behavior-related questions. The results are shown in Tab.~\ref{table:behavior}.

\subsection{Improving Human Behavior Understanding}
\vspace{-2mm}

In this section, we show that \ours~ can facilitate versatile tasks of human behavior understanding. By finetuning different baseline models on our dataset, we observe significant performance gains. 

\begin{figure*}[t]
    \centering
 \includegraphics[width=0.98\linewidth]{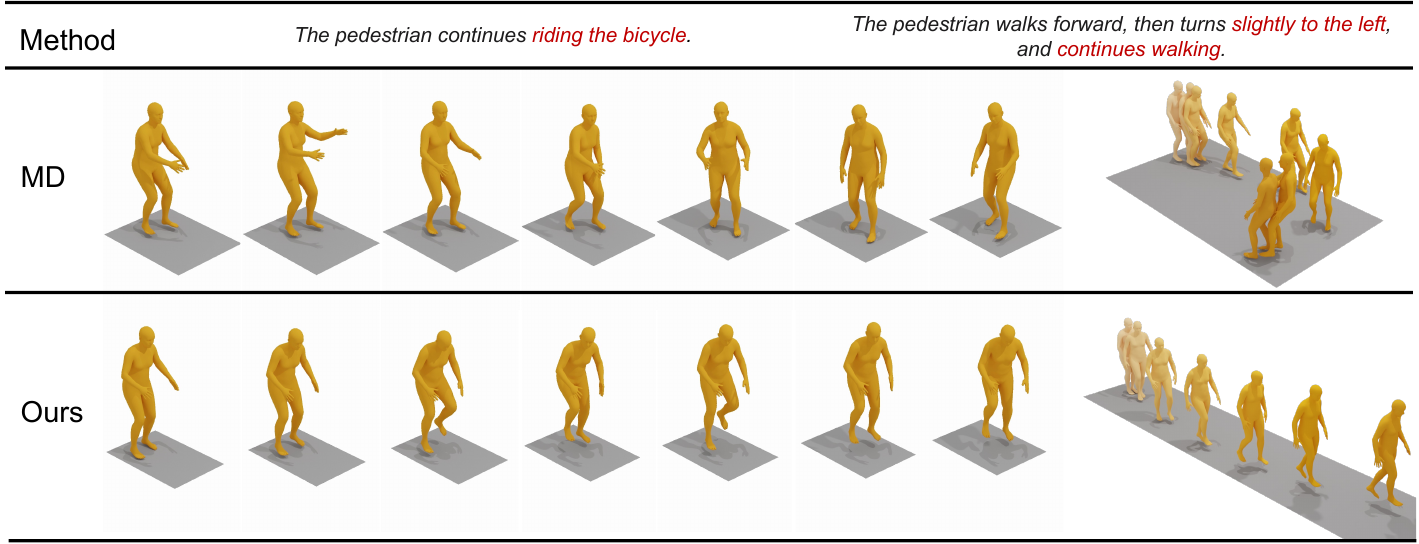}
    \caption{\textbf{Qualitative comparison of Motion Generation.} 
    The baseline model (MotionDiffuse, MD) is not capable to generate proper motions in driving scenes. After fine-tuning on \ours (second row), the model  demonstrates the ability to generate human motions in autonomous driving scenarios.
    }
    \label{fig:motion_generation_comparsion}
\vspace{-3mm}
\end{figure*}

\vspace{-2mm}
\paragraph{Motion Prediction.}
We trained PhysMoP~\cite{zhang2024incorporating} on the mixed data of 3DPW~\cite{vonMarcard2018} and \ours, and compared it with that trained on 3DPW only. We then evaluate both variants on the original 3DPW dataset, following the original settings of PhysMoP. As shown in Tab.~\ref{tab:bene_motion}, the model training with our \ours~data generalizes to 3DPW and significantly outperforms the baseline model by 9.49 average MPJPE and 1.1 ACCL. 
\vspace{-2mm}
\paragraph{Motion Generation.} We show that fine-tuning on \ours~can narrow the domain gap between generic text-to-motion generation to motions in driving scenes. We fine-tune MotionDiffuse~\cite{zhang2024motiondiffuse} on \ours, and we observe a significant improvement on FID, as shown in Tab.~\ref{tab:bene_motion_gen}. The visualization results in Fig.~\ref{fig:motion_generation_comparsion} further show the effectiveness of fine-tuning on our dataset. 
\vspace{-2mm}
\paragraph{Behavior VQA.}
We employed QWen2.5-VL~\cite{bai2023qwen} as our baseline. We fine-tuned the baseline model on our dataset and evaluated both models using the average accuracy and the F1 score on \ours-T. As shown in Tab.~\ref{tab:bene_vqa}, the fine-tuned model achieves a significant performance gain of 15.96\% and 15.19\% with respect to the average precision and F-1 score. 
\vspace{-2mm}
\paragraph{Intention Prediction:} Intention prediction answers the question of whether or not a person is going to cross the street, which is a special aspect of behavior QA. We select TrEP~\cite{zhang2023trep} as the baseline. 
We train the baseline model on the JAAD~\cite{kotseruba2016joint} dataset, which is a widely used dataset for intention prediction. We then compared the baseline model to that trained on the mixture of JAAD~\cite{kotseruba2016joint} and \ours. We observe \ours~significantly contribution to a performance gain of the baseline model. All evaluations are conducted on the JAAD~\cite{jiang2024mantis} test set following the original settings of TrEP~\cite{zhang2023trep}.

\begin{table}[!htb]
    \begin{minipage}[t]{.32\linewidth}
          \renewcommand{\arraystretch}{1.0}

      \caption{\small \textbf{Benefiting Motion Prediction.} We train the PhysMoP~\cite{zhang2024incorporating} on a mixed set of \ours~ and 3DPW~\cite{vonMarcard2018} and compared with that trained on 3DPW only. The evaluation is conducted on 3DPW, following the original setting.}
      
      \label{tab:bene_motion}
      \small
      \centering
      \resizebox{\textwidth}{!}{
        \begin{tabular}{ccc}
\toprule
Train Set                                                   & MPJPE-avg$\downarrow$           & ACCL$\downarrow$                \\
\midrule
3DPW                                                         &        47.67             &          3.8      \\
3DPW+\ours       & 38.18 & 2.7 \\

\bottomrule

\end{tabular}
}
    \end{minipage}%
    \hfill
    \begin{minipage}[t]{.32\linewidth}
      \centering
            \renewcommand{\arraystretch}{0.85}
\vspace{-3pt}
        \caption{\small\textbf{Benefiting Intention Prediction.} We train the baseline TrEP~\cite{zhang2023trep} with a mixture of JAAD~\cite{kotseruba2016joint} and \ours. We then evaluate on the JAAD dataset. Compared to training on JAAD alone, training with our dataset performs significantly better.}
        \label{tab:bene_itention}
      \resizebox{\textwidth}{!}{
        \begin{tabular}{cccc}
    \toprule
    Train Set & Accuracy $\uparrow$ & F1-score$\uparrow$ & AuROC$\uparrow$ \\
    \midrule
    JAAD & 84.49 & 84.45 & 92.98\\
    JAAD+\ours & 91.89 & 91.89 & 97.72 \\
  \bottomrule
\end{tabular}
}
    \end{minipage} 
\hfill
    \begin{minipage}[t]{.32\linewidth}
      \centering
            \renewcommand{\arraystretch}{0.95}

        \caption{\small\textbf{Benefiting Motion Generation.} 
        Models fine-tuned on \ours ~(labeled with *) can better generate motions in driving domains and achieves notably lower FID.
        }
        \label{tab:bene_motion_gen}
      \resizebox{\textwidth}{!}{
        \begin{tabular}{ccc}
    \toprule
    Model & FID$\downarrow$ & Multi Modality \\
    \midrule
    Real & 0.0020 & - \\
    MotionDiffuse & 39.27 & 2.36 \\
    MotionDiffuse* & 1.86 & 2.31  \\
    MotionGPT & $27.06$ & $5.42$ \\
MotionGPT* & $8.44$ & $3.77$ \\
    
  \bottomrule
\end{tabular}
}
    \end{minipage} 
\end{table}

\section{Conclusion and Limitation}
\vspace{-2mm}
\label{sec:conc}
\paragraph{Conclusion.} In this work, we present ~\ours, a large-scale dataset for human behavior understanding. The \bfours~ dataset consists of 57k human instances, each of them are richly annotated with 3D motion sequences, text descriptions, and labeling of 13 critical behaviors. We collected driving videos from various sources and developed a human-in-the-loop annotation pipeline to get high-quality annotations. We evaluated and analyzed the baseline models on the proposed dataset regarding the task of motion prediction, text-to-motion generation, and human behavior VQA. We also conduct experiments to show how \ours~can benefit these tasks.

\vspace{-2mm}
\paragraph{Limitation.} \ours~provided a comprehensive dataset for human behavior understanding and developed an annotation pipeline that involves minimal human effort. The capability of VLMs to understand human behavior limits the development of a fully automatic annotation pipeline.

\vspace{-2mm}

{
    \small
    \bibliographystyle{unsrt}
    \bibliography{main}
}

\appendix

\newpage
\section*{Appendices}

We describe the details of the datasets in Sec.~\ref{sec:sup-data}, including the details of the data collection, human annotation, and the visualization results from the dataset. We then describe the details of the experiments we conduct in Sec.~\ref{sec:sup-exp}.

\section{Dataset Details}
\label{sec:sup-data}
\subsection{Data Collection}
For the YouTube source, we filtered and downloaded videos with the \emph{Creative Commons Attribution license (reuse allowed)} license using yt-dlp~\cite{ytdlp}. We also filtered the videos by checking whether they contain the keywords ``drive'' or ``driving'' in their title. We collect video clips that sum to over 11 hours, consisting of 10k human instances.
We list the details of used YouTube channels in Tab.~\ref{tab:supp-youtube}.

\begin{table}[ht]
\centering
\caption{\textbf{Collected YouTube Raw Data}}
\label{tab:supp-youtube}
\begin{tabular}{cccc}
\toprule
Channel Name & License & Collected Minutes & Human Instances \\
\midrule
TravelRelaxListen &    
Creative Commons Attribution      &      179           &       2028          \\
planetearthtraveler &    Creative Commons Attribution     &     92           &         3122        \\
RoamingBrit     & Creative Commons Attribution    &      65     &     497        \\
Evan-Explores       & Creative Commons Attribution      &    147       &       62       \\
VietnamSilentRoutes       & Creative Commons Attribution       &    187       &       4812       \\
\bottomrule
\end{tabular}
\end{table}

\subsection{Human Annotation}
\begin{figure*}[t]
    \centering
 \includegraphics[width=0.95\linewidth]{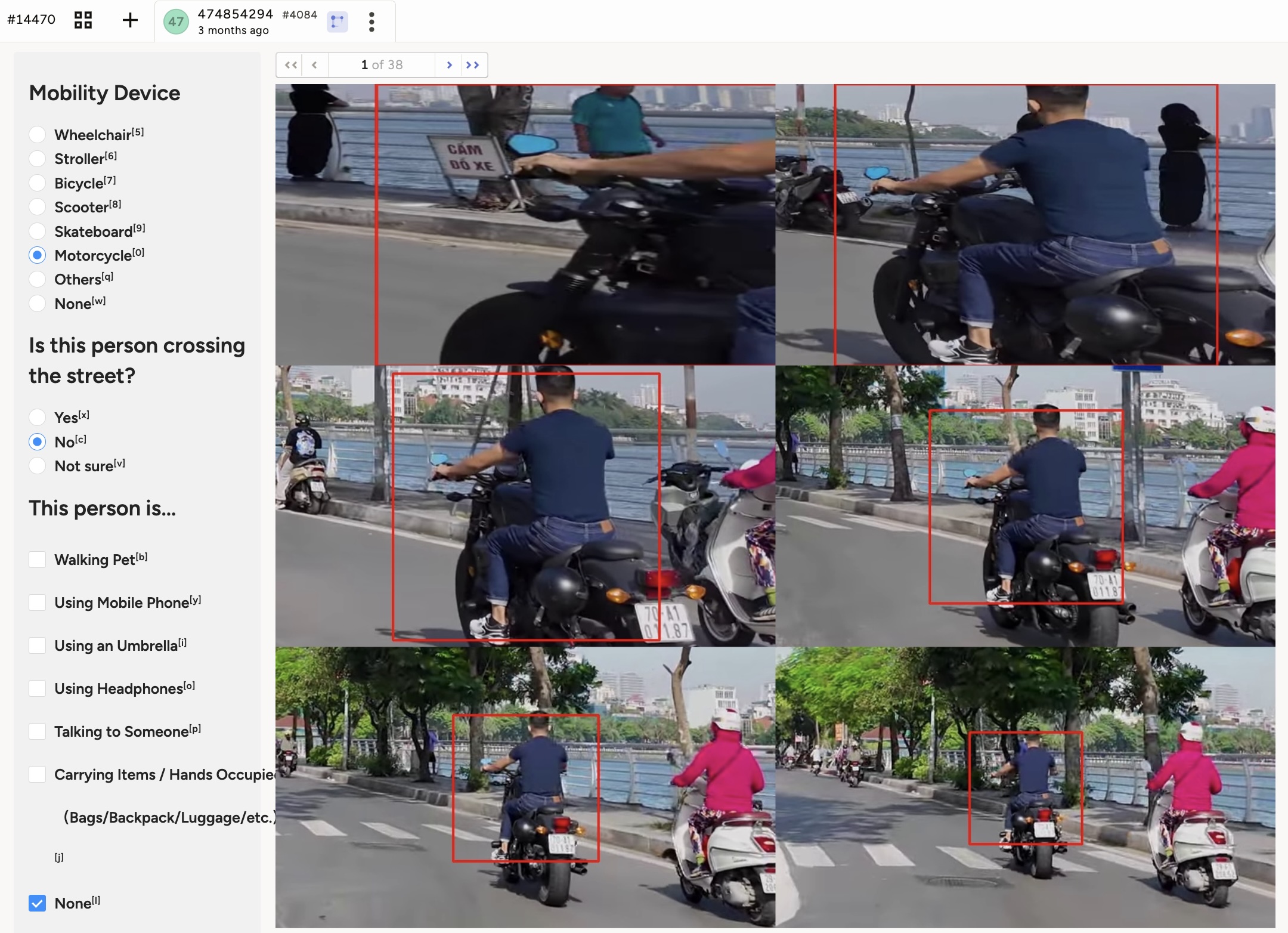}
    \caption{\textbf{Annotation Interface.} 
    }
    \label{fig:supp-interface}
\vspace{-3mm}
\end{figure*}

We present the details of the human annotators for the \ours-H and \ours-T subsets. We employ 12 annotators. Each annotator is provided with 6 frames of the video clip. The target human is cropped and labeled with a bounding box. We asked the annotators to check for each behavior that the human holds. We pay the annotator 0.15 CNY per video clip, leading to roughly 50 CNY per hour. We use label studio~\cite{LabelStudio} to build the annotation environment. The English version of the annotation interface is shown in Fig.~\ref{fig:supp-interface}.

\subsection{Details of Hierarchical Text Annotation}
To bridge the gap between parameterized SMPL motion and semantic-level understanding, we implement a hierarchical text annotation pipeline that translates joint-level motion into structured language descriptions.

We use PoseScript~\cite{delmas2022posescript} to extract joint-wise behavioral descriptions (the low-level description) from SMPL pose sequences. Unlike the original method, which outputs a single paragraph per frame summarizing the full-body posture, our approach decomposes captions into joint-specific phrases by isolating text segments corresponding to individual joints. We focus on the most salient joint movements and retain only those joints mentioned in at least 50\% of the frames within a sequence. To ensure temporal consistency, we align the low-level descriptions of each selected joint across the sequence. These temporally aligned joint descriptions are aggregated using a large language model to produce concise summaries of each joint’s motion over time. The low-level descriptions are utilized to support the subsequent aggregation of high-level semantic behavioral descriptions. We show an example of the low-level (joint-wise) descriptions in Box~1.

\begin{tcolorbox}[
label=box:supp-joint_and_low_level_descriptions,  
colback=gray!5!white, 
colframe=gray!75!black, 
 title=Box 1. Example of Low-Level Descriptions,
]
    \textbf{Selected Key Frames for Reference:}
    \begin{center}
        \includegraphics[width=0.9\linewidth]{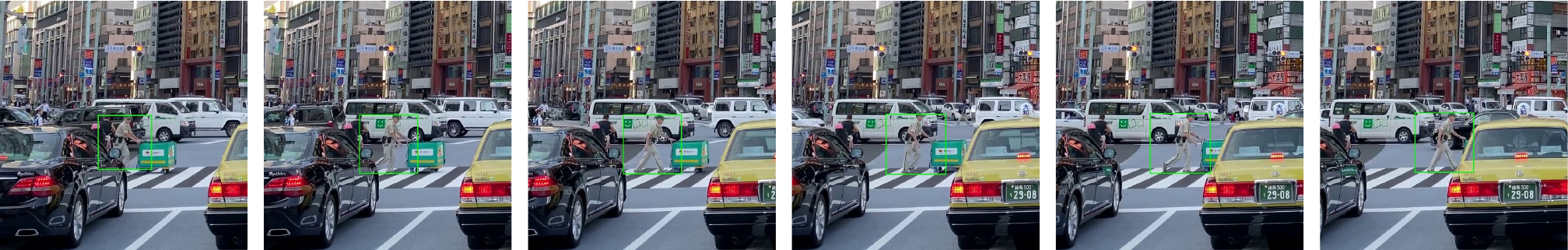}
    \end{center}
    
    \vspace{-2pt}
    \textbf{Low-Level (Joint-wise) Descriptions:}
    \begin{itemize}[leftmargin=2em]
        \item \textbf{Elbow:} The elbows are generally bent throughout the sequence, with some variation in the degree of bending.
        \item \textbf{Forearm:} The forearm remains mostly horizontal with occasional alignment with the thighs and shins, indicating a stable position with some minor adjustments.
        \item \textbf{Knee:} The knees are generally bent with slight variations, occasionally straightening or separating at shoulder width, indicating a dynamic posture.
        \item \textbf{Left Elbow:} The left elbow remains mostly bent throughout the sequence, with occasional variations indicating slight changes in its position.
        \item ...
    \end{itemize}

\end{tcolorbox}

Then a vision-language model is employed to aggregate the low-level (joint-wise) descriptions with the high-level semantic behavior information from the keyframes. The instuctions for this procedure is demonstrated in Box~2.

\begin{tcolorbox}[
label=box:supp-text_description_template,  
colback=gray!5!white, 
colframe=gray!75!black, 
title=Box 2. Instruction Template for Aggregated Text Description,
]
    \textbf{System Prompt:} You are an expert in human motion analysis and natural language refinement.\\
    Your task is to reorganize and clarify human pose descriptions derived from SMPL-based 3D models. Ensure all descriptions are structured, concise, and consistent, while strictly preserving the original semantics—do not add, remove, or alter key motion elements.\\
    When provided with low-level pose information (e.g., joint-based summaries or posecodes), use it to enhance the clarity or specificity of the description, especially regarding posture and limb motion. However, maintain overall cohesion and naturalness, with a clear focus on the pedestrian’s observable behavior in context.\\
    
    \textbf{User Prompt:} You are given a sequence of images showing a pedestrian in a green bounding box, captured by a front-facing car camera, along with a dictionary of low-level joint-based pose descriptions.\\
    Your task is to:
    \begin{itemize}[leftmargin=2em]
      \item Write a concise, one- to two-sentence summary capturing the pedestrian’s overall motion and any key changes in action throughout the sequence.
      \item Emphasize how the pedestrian’s behavior evolves over time (e.g., walking, stopping, turning, crouching, changing direction).
      \item Use the low-level pose descriptions to refine details when relevant, but avoid excessive mechanical phrasing or redundancy.
      \item Describe only what is clearly supported by the visual and pose data — do not speculate.
    \end{itemize}
    
    Format your final response using the following tags:
    \texttt{<description>} human pose description here. \texttt{<description\_end>}\\
    Example:
    \texttt{<description>} The pedestrian stepped forward with arms raised, paused briefly, then lowered their body and turned slightly to the left. \texttt{<description\_end>}\\
    
    Here is the low-level pose description: \{\textcolor{blue}{Input Low-Level Description}\}

\end{tcolorbox}

\section{Experiments}
\label{sec:sup-exp}
\vspace{-2mm}
\subsection{Human Behavior VQA Details.}
\vspace{-2mm}
For each behavior, we construct closed-ended questions based on the template. We provided 4 to 6 frames to the VLMs. The frames are uniformly sampled from the video clips. 
To reduce randomness, we will ask the VLMs twice, one direct question and one counter question respectively. The VLM should answer both questions correctly in order to be evaluated as correct on the behavior. We show two example in Box~3
and Box~4 
(both answered correctly). We find some VLMs are hard to follow the instructions, thus if the answer has the wrong format, we will roll back at most three times and repeat the same question. An example is shown in Box~5.

\subsection{Additional Results on Motion Generation}
To better show the performance improvement of motion generation tasks after finetuning on \ours~dataset, we provide some videos of generated motions along with the supplementary materials. We provided three extra cases to show the fine-tuned model is more capable of generating human motion in the driving scene. The video is named following ``<casenumber>-<Baseline/Finetune>-<behavior\_label>-<Mesh/Skeleton>". The first case shows that the baseline is not capable of generating feasible motion of riding a bike. In the second case, the baseline model only captured the ''walking" keyword and ignores the behavior of ''talking to someone", while the fine-tuned model can generate a more plausible motion with hand postures. In the Third case, we add a more detailed description of human pose, and the fine-tuned model shows the ability to follow the more detailed instructions.

\begin{tcolorbox}[
label={box_supp_vqa_carrying_item},  
colback=gray!5!white, 
colframe=gray!75!black, 
title=Box 3. Example of Human Behavior VQA: Carrying Items,,
]
    \textbf{Frames:}
    \begin{center}
        \includegraphics[width=0.9\linewidth]{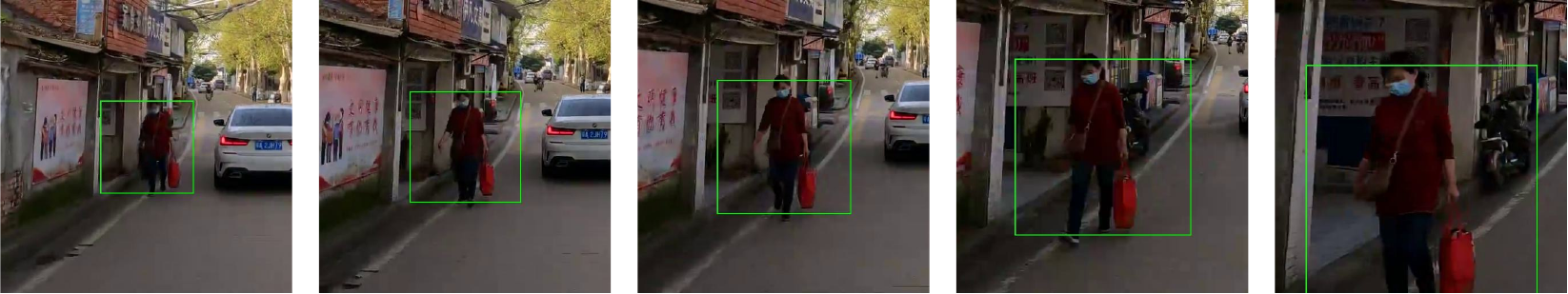}
    \end{center}
    
    \vspace{0.5em}
    \textbf{System Prompt:} You are an expert pedestrian behavior labeler, specializing in analyzing pedestrians' behavior on the road. You will perform visual analysis on multiple sequential images over time.
    
    \vspace{0.5em}
    \textbf{Question 1:} The images show a pedestrian in a bounding box with surrounding context. The image is a cropped frame from a car's front camera. \\
    Is the pedestrian \textcolor{red}{carrying items}? \\
    Please answer y or n, only one letter.\\
    
    \textbf{Answer:} \texttt{y}
    
    \vspace{0.5em}\noindent\rule{\linewidth}{0.5pt}\vspace{0.5em}
    
    \textbf{Question 2:} The images show a pedestrian in a bounding box with surrounding context. The image is a cropped frame from a car's front camera. \\
    The pedestrian is \textcolor{red}{not carrying items}. Is this statement correct? \\
    Please answer y or n, only one letter.\\
    
    \textbf{Answer:} \texttt{n}
\end{tcolorbox}

\begin{tcolorbox}[
label={box_supp_vqa_using_scooter},  
colback=gray!5!white, 
colframe=gray!75!black, 
title=Box 4. Example of Human Behavior VQA: Using Scooter,
]
    \textbf{Frames:}
    \begin{center}
        \includegraphics[width=0.9\linewidth]{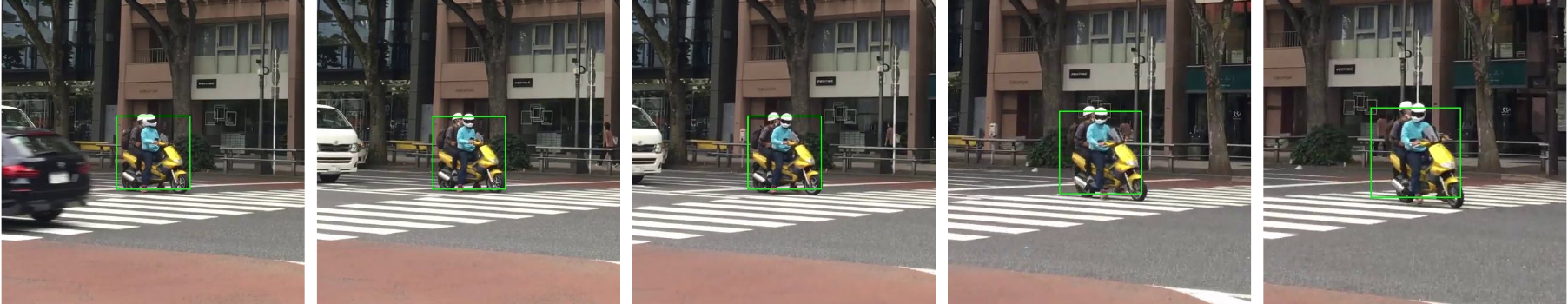}
    \end{center}
    
    \vspace{0.5em}
    \textbf{System Prompt:} You are an expert pedestrian behavior labeler, specializing in analyzing pedestrians' behavior on the road. You will perform visual analysis on multiple sequential images over time.
    
    \vspace{0.5em}
    \textbf{Question 1:} The images show a pedestrian in a bounding box with surrounding context. The image is a cropped frame from a car's front camera. \\
    Is the pedestrian \textcolor{red}{using a scooter}? \\
    Please answer y or n, only one letter.\\
    
    \textbf{Answer:} \texttt{y}
    
    \vspace{0.5em}\noindent\rule{\linewidth}{0.5pt}\vspace{0.5em}
    
    \textbf{Question 2:} The images show a pedestrian in a bounding box with surrounding context. The image is a cropped frame from a car's front camera. \\
    The pedestrian is \textcolor{red}{not using a scooter}. Is this statement correct? \\
    Please answer y or n, only one letter.\\
    
    \textbf{Answer:} \texttt{n}
\end{tcolorbox}

\begin{tcolorbox}[
label=box:supp-vqa_using_phone,  
colback=gray!5!white, 
colframe=gray!75!black, 
title=Box 5. Example of Human Behavior VQA: Using a Phone,
]
    \textbf{Frames:}
    \begin{center}
        \includegraphics[width=0.9\linewidth]{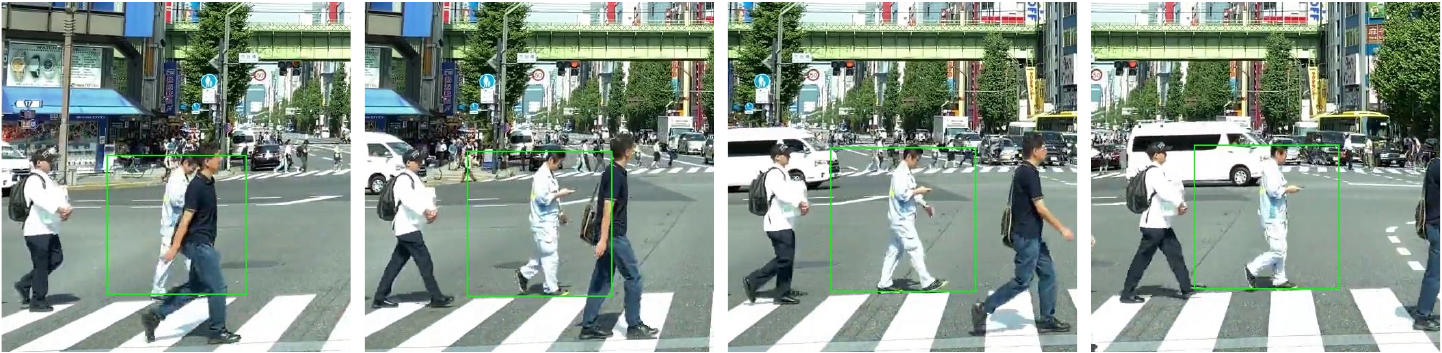}
    \end{center}
    
    \vspace{0.5em}
    \textbf{System Prompt:} You are an expert pedestrian behavior labeler, specializing in analyzing pedestrians' behavior on the road. You will perform visual analysis on multiple sequential images over time.
    
    \vspace{0.5em}
    \textbf{Question 1:} The images show a pedestrian in a bounding box with surrounding context. The image is a cropped frame from a car's front camera. \\
    Is the pedestrian \textcolor{red}{using a phone}? \\
    Please answer y or n, only one letter.\\
    
    \textbf{Answer:} \texttt{This person is crossing the street.}

    \vspace{0.5em}
    \textbf{[RollBack] Question 1:} The images show a pedestrian in a bounding box with surrounding context. The image is a cropped frame from a car's front camera. \\
    Is the pedestrian \textcolor{red}{using a phone}? \\
    Please answer y or n, only one letter.\\
    
    \textbf{[RollBack] Answer:} \texttt{y}
    
    \vspace{0.5em}\noindent\rule{\linewidth}{0.5pt}\vspace{0.5em}
    
    \textbf{Question 2:} The images show a pedestrian in a bounding box with surrounding context. The image is a cropped frame from a car's front camera. \\
    The pedestrian is \textcolor{red}{not using a phone}. Is this statement correct? \\
    Please answer y or n, only one letter.\\
    
    \textbf{Answer:} \texttt{n}
    
\end{tcolorbox}

\begin{table}[ht]
\centering
\caption{\textbf{Additional Results on Motion Generation}}
\label{tab:supp-motion-gen}
\begin{tabular}{ccc}
\toprule
Model & FID$\downarrow$ & Multi Modality \\
\midrule
Real & 0.0020 & - \\
MotionGPT Baseline & $27.06^{\text{\scriptsize$\pm$0.645}}$ & $5.42^{\text{\scriptsize$\pm$0.124}}$ \\
MotionGPT Finetuned & $8.44^{\text{\scriptsize$\pm$0.203}}$ & $3.77^{\text{\scriptsize$\pm$0.096}}$ \\
\bottomrule
\end{tabular}
\end{table}

\subsection{Setting Details in Finetuning Models}
We provide the details of the fine-tuning procedure used in Sec. 5.4 of the main paper.

\paragraph{Motion Prediction.}The training data consists of the entire 3DPW dataset (training set)~\cite{vonMarcard2018} and 30k randomly selected human sequences from \ours. All training settings follow the default configurations described in these papers. Since the frame rate of 3DPW is 25 FPS, while ours is 10 FPS, we upsample all sequences in our dataset to 50 FPS and then downsample them to 25 FPS. Training is conducted on a NVIDIA RTX 4090 GPU.

\paragraph{Motion Generation.}We select the test set from our MMHU dataset and split it into training, validation, and test subsets with a 7:1:2 ratio for this experiment. The main training settings and model architecture follow the default configurations in MotionDiffuse~\cite{zhang2024motiondiffuse}. We use the Adam optimizer with a learning rate of 0.0002. For fine-tuning, we set the batch size to 192 and train for 20 epochs using a single NVIDIA RTX 6000 Ada GPU.

\paragraph{Human Behavior VQA.}We apply LoRA ~\cite{hu2022lora} fine-tuning to Qwen2.5-3B-Instruct by using LLaMA-Factory Framework~\cite{zheng2024llamafactory}. The visual branch is frozen, and LoRA is applied to all other MLP layers of the model. Our LoRA settings are: lora\_rank = 8 and lora\_alpha = 16. The model is trained for one epoch on a subset of the MMHU dataset and evaluated on the MMHU-T dataset. Due to class imbalance, we employed re-sampling to make sure the same training data from each behavior class.
The training is conducted on 4 NVIDIA RTX 6000 Ada GPUs with a total batch size of 128. The learning rate is set to 1e-4, with a warmup ratio of 0.1, and the learning rate scheduler is CosineAnnealingLR. All input images are cropped to $256 \times 256$ around the bounding box, or resized to $256 \times 256$ if the bounding box is larger than $256 \times 256$.

\paragraph{Intention Prediction.}The training data consists of the JAAD dataset and \ours~training set. All training settings primarily follow the default configurations in Trep~\cite{zhang2023trep} on the JAAD dataset. All models are implemented using the Adam optimizer with a learning rate of 0.005. We adopt an early stopping strategy with a patience of 5, and experiments are conducted on a single NVIDIA RTX 6000 Ada GPU.

\end{document}